\documentclass[5pt]{article}

\usepackage[utf8]{inputenc}
\usepackage[T1]{fontenc}
\usepackage{lmodern}

\usepackage[a4paper,margin=1in]{geometry}

\usepackage{graphicx}
\usepackage{caption}
\usepackage{subcaption}

\usepackage{amsmath,amssymb,amsfonts}
\usepackage{bm} 
\DeclareMathOperator*{\argmax}{arg\,max}
\DeclareMathOperator*{\argmin}{arg\,min}

\usepackage{algorithmic}

\usepackage{pgf}
\usepackage{pgfmath}
\usepackage{pgfplots}
\pgfplotsset{compat=1.18}
\usepackage{pgffor}
\usepackage{tikz}

\usepackage{tabularray}
\usepackage{multirow, makecell, booktabs}
\usepackage[table]{xcolor}
\usepackage{collcell}
\usepackage{tabularx}

\usepackage{url}
\usepackage[hidelinks]{hyperref}

\newcommand{\cvconfplot}[4]{%
  \setlength{\tabcolsep}{1pt}%
  \renewcommand{\arraystretch}{1.6}%
  \raisebox{-0.5\height}{%
    \pgfmathsetmacro{\valA}{#1/100}%
    \pgfmathsetmacro{\valB}{#2/100}%
    \pgfmathsetmacro{\valC}{#3/100}%
    \pgfmathsetmacro{\valD}{#4/100}%
    \pgfmathsetmacro{\grayA}{max(0,min(1,1-0.5*(\valA-0.5)))}%
    \pgfmathsetmacro{\grayB}{max(0,min(1,1-0.5*(\valB-0.5)))}%
    \pgfmathsetmacro{\grayC}{max(0,min(1,1-0.5*(\valC-0.5)))}%
    \pgfmathsetmacro{\grayD}{max(0,min(1,1-0.5*(\valD-0.5)))}%
    \begin{tabular}{c|c c}
      & \scriptsize \(z_1\) & \scriptsize \(z_2\) \\ \hline
      \scriptsize \(y_1\) & \cellcolor[gray]{\grayA}\scriptsize #1 & \cellcolor[gray]{\grayB}\scriptsize #2 \\
      \scriptsize \(y_2\) & \cellcolor[gray]{\grayC}\scriptsize #3 & \cellcolor[gray]{\grayD}\scriptsize #4
    \end{tabular}%
  }%
}
\newcommand{\cvconfplotone}[2]{%
  \setlength{\tabcolsep}{1pt}%
  \renewcommand{\arraystretch}{1.6}%
  \raisebox{-0.5\height}{%
    \pgfmathsetmacro{\valA}{#1/100}%
    \pgfmathsetmacro{\valC}{#2/100}%
    \pgfmathsetmacro{\grayA}{max(0,min(1,1-0.5*(\valA-0.5)))}%
    \pgfmathsetmacro{\grayC}{max(0,min(1,1-0.5*(\valC-0.5)))}%
    \begin{tabular}{c|c}
      & \scriptsize \(z_1\) \\ \hline
      \scriptsize \(y_1\) & \cellcolor[gray]{\grayA}\scriptsize #1 \\
      \scriptsize \(y_2\) & \cellcolor[gray]{\grayC}\scriptsize #2
    \end{tabular}%
  }%
}

\usepackage[numbers]{natbib}  

\usepackage{titlesec}
\titleformat{\section}{\large\bfseries}{\thesection}{1em}{}
\titleformat{\subsection}{\normalsize\bfseries}{\thesubsection}{1em}{}

\title{Mitigating Shortcut Learning via Feature Disentanglement in Medical Imaging: A Benchmark Study}

\author{
Sarah M\"uller\textsuperscript{1,2}\thanks{ Corresponding author: \texttt{sar.mueller@uni-tuebingen.de}} 
\and
Philipp Berens\textsuperscript{1,2}
}

\date{}

\begin{document}
\maketitle

\noindent
\textsuperscript{1} Hertie Institute for AI in Brain Health, Faculty of Medicine, University of Tübingen, Germany\\
\textsuperscript{2} Tübingen AI Center, University of Tübingen, Germany

\begin{abstract}
Although deep learning models in medical imaging often achieve excellent classification performance, they can rely on shortcut learning, exploiting spurious correlations or confounding factors that are not causally related to the target task. This poses risks in clinical settings, where models must generalize across institutions, populations, and acquisition conditions. Feature disentanglement is a promising approach to mitigate shortcut learning by separating task-relevant information from confounder-related features in latent representations.
In this study, we systematically evaluated feature disentanglement methods for mitigating shortcuts in medical imaging, including adversarial learning and latent space splitting based on dependence minimization. We assessed classification performance and disentanglement quality using latent space analyses across one artificial and two medical datasets with natural and synthetic confounders. We also examined robustness under varying levels of confounding and compared computational efficiency across methods. We found that shortcut mitigation methods improved classification performance under strong spurious correlations during training. Latent space analyses revealed differences in representation quality not captured by classification metrics, highlighting the strengths and limitations of each method. Model reliance on shortcuts depended on the degree of confounding in the training data. The best-performing models combine data-centric rebalancing with model-centric disentanglement, achieving stronger and more robust shortcut mitigation than rebalancing alone while maintaining similar computational efficiency. The project code is publicly available:~\url{https://github.com/berenslab/medical-shortcut-mitigation}.
\end{abstract}

\section{Introduction}
Deep learning models have achieved remarkable performance in medical imaging tasks such as disease classification, segmentation, and prognosis prediction. However, growing evidence suggests that models sometimes rely on \emph{shortcut learning}, i.e., they exploit spurious correlations or confounding factors that are predictive in the training data but not causally related to the target task~\cite{geirhos2020shortcut, steinmann2024navigating, wang2024effect}. In the machine learning literature, this phenomenon is described using various terms, including shortcuts, spurious correlations, Clever Hans behavior, or confounders~\cite{steinmann2024navigating}. The impact of shortcut learning is particularly critical in the medical domain, where models are expected to generalize across institutions, populations, and medical devices. In medical imaging, for example, such shortcuts may arise from acquisition protocols, scanner-specific artifacts, demographic imbalances, or hospital-specific workflows~\cite{degrave2021ai,yang2022machine,chen2025international,singh2022generalizability,kitamura2020retraining,wang2020inconsistent,ahluwalia2023subgroup}, leading to fragile models that fail under distribution shifts and raise concerns about safety and trustworthiness.

\begin{figure*}[htbp]
    \centering
    \includegraphics[width=\linewidth]{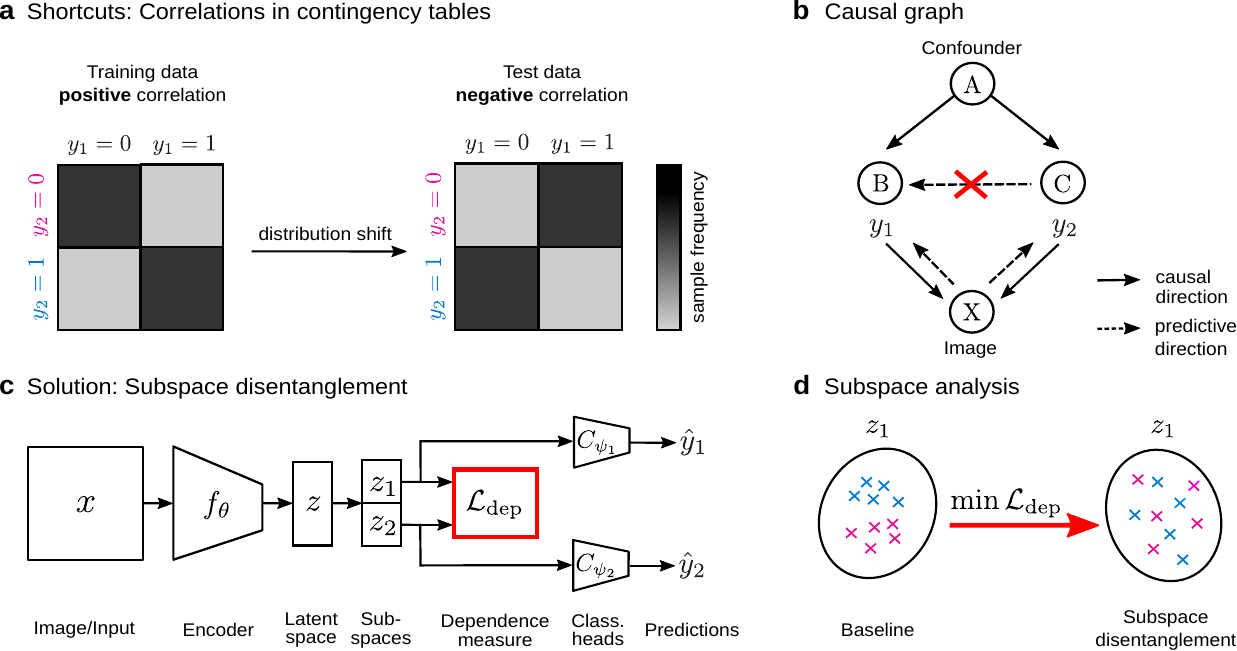}
    \caption{Overview of shortcut learning and mitigation via feature disentanglement. \textbf{a} Example of a spurious correlation between two binary tasks in the training data that reverses at test time, illustrating a distribution shift. \textbf{b} Causal graph in which a confounder affects both tasks \(y_1\) and \(y_2\), while the image \(X\) is generated from both; the predictive association from \(y_2\) to \(y_1\) constitutes a shortcut. \textbf{c} Subspace disentanglement architecture that splits the latent representation into task-specific subspaces and minimizes their statistical dependence by minimizing a dependence measure \(\mathcal{L}_\text{dep}\). \textbf{d} Comparison of latent representations showing that disentanglement reduces clustering of the confounding label \(y_2\) in the target-task subspace \(z_1\).}
    \label{fig:overview-figure}
\end{figure*}

For example, models trained to detect medical pathologies like disease features may inadvertently rely on hospital-specific markers, image resolution differences, or demographic attributes correlated with disease prevalence rather than true pathological features (Fig.\,\ref{fig:overview-figure}\,\textbf{a}). Such behavior not only negatively affects out-of-distribution performance~\cite{koh2021wilds,koch2023subgroup} but may also lead to unfair decision-making, because shortcut learning often manifests as reliance on sensitive attributes~\cite{brown2023detecting}. As emerging regulatory and ethical frameworks for medical AI, such as the EU AI Act~\cite{EU_AI_Act_overview}, WHO guidelines for trustworthy AI in health~\cite{WHO_AI_health_ethics}, and FDA oversight of AI-enabled medical devices~\cite{FDA_AI_SaMD_overview} stress robustness, fairness, and transparency, approaches for mitigating shortcut learning are becoming increasingly important. From a methodological perspective, shortcut learning can be interpreted through a causal lens, where models capture correlations induced by confounders rather than causal mechanisms underlying the target variable (Fig.~\ref{fig:overview-figure}\textbf{b})~\cite{castro2020causality,mueller2024disent}. These perspectives jointly motivate approaches that explicitly separate task-relevant information from spurious or confounding factors.

Prior work on mitigating shortcut learning can be broadly categorized into dataset-centric and model-centric approaches. Dataset-centric methods aim to reduce spurious correlations through data curation~\cite{ahmed2022achieving, portal2022new}, rebalancing~\cite{li2019repair, idrissi2022simple, sagawa2020investigation}, augmentation~\cite{seo2022information, lee2021learning, anders2022finding}, or synthetic confounder control~\cite{teso2019explanatory, plumb2021finding, nauta2021uncovering, struppek2023leveraging, kwon2024learning, noohdani2024decompose, mao2021generative}. Model-centric approaches mitigate shortcut learning by modifying training objectives or architectures, for example through explanation-based regularization~\cite{ross2017right, schramowski2020making}, adversarial objectives that enforce invariance to confounders~\cite{ganin2015unsupervised, kamnitsas2017unsupervised}, or invariant risk minimization frameworks that seek predictors stable across environments~\cite{arjovsky2019invariant, zare2022removal, chen2022does, yang2024identifying}.

Within this space, \emph{feature disentanglement} has emerged as a promising paradigm in which latent representations are explicitly decomposed into task-relevant and confounder-related subspaces~\cite{fay2023avoid, mueller2024disent}. Such decompositions have been explored using generative models~\cite{louizos2015variational, xie2020mi, yang2022chroma} and encoder-based classifiers~\cite{fay2023avoid, mueller2025bench, wang2024navigate}, and are typically enforced via statistical dependence measures, including mutual information minimization~\cite{xie2020mi, fay2023avoid}, distance correlation~\cite{sz2007dCor, mueller2024disent}, or kernel-based criteria such as maximum mean discrepancy~\cite{gretton2012kernel}. These methods reduce shortcut reliance by separating confounding factors from predictive signal (Fig.~\ref{fig:overview-figure}\textbf{c,d}).

Building on preliminary work~\cite{mueller2025bench}, the present study addresses open questions in the evaluation of shortcut mitigation methods. In particular, existing studies typically consider individual approaches or limited experimental settings, leaving the comparative strengths, limitations, and robustness of different methods insufficiently understood. We therefore conduct a systematic and large-scale evaluation of feature disentanglement approaches for shortcut mitigation: First, we provide a comprehensive comparison of feature disentanglement methods for mitigating shortcut learning, evaluated in terms of classification performance, disentanglement quality, robustness across correlation strengths, and computational cost. Second, we analyze consistent trends across three publicly available datasets, including a controlled toy dataset and two medical imaging datasets with both natural and synthetic confounders. Third, we demonstrate that combining data-centric interventions with model-centric latent space disentanglement yields more effective and robust mitigation of shortcut learning than either approach alone.

\section{Related Work}
A wide range of approaches has been proposed to mitigate shortcut learning, which can be broadly categorized into data-centric and model-centric techniques.

\subsection{Data-centric techniques}
Data-centric approaches aim to reduce spurious correlations by modifying the training data or by a data sampling strategy. A common sampling strategy is resampling, where samples are over- or under-sampled to weaken correlations between target labels and spurious attributes~\cite{li2019repair, idrissi2022simple, sagawa2020investigation}. Closely related are sample reweighting methods, which adjust the contribution of individual samples during training, for example through weighted softmax objectives~\cite{luo2022pseudo} or by fine-tuning models previously trained on biased data using reweighted samples~\cite{liu2021just, kirichenko2022last, izmailov2022feature}.

Another line of work focuses on data curation, where datasets are manually or semi-automatically refined to reduce confounding effects~\cite{ahmed2022achieving, portal2022new}. Static data augmentation methods aim to explicitly manipulate spurious features, for example by segmenting and masking shortcut-related regions~\cite{teso2019explanatory, plumb2021finding, nauta2021uncovering, struppek2023leveraging}, synthesizing new samples to balance target and spurious factors~\cite{kwon2024learning, noohdani2024decompose, mao2021generative}, or applying generic augmentation strategies such as MixUp~\cite{zhang2017mixup, yao2022improving, wu2023discover} and CutMix~\cite{yun2019cutmix}. 

Adaptive data augmentation methods modify the training data dynamically during optimization, for instance by injecting stochastic label noise~\cite{seo2022information}, swapping shortcut features between samples during training~\cite{lee2021learning}. Other approaches augment latent representations at intermediate network layers, guided by local explainability methods such as Layer-wise Relevance Propagation (LRP), which are used to identify shortcut-related features and regions in the model’s representations~\cite{anders2022finding}. Finally, feature reweighting approaches such as invariant risk minimization (IRM)~\cite{arjovsky2019invariant} are based on the assumption that the causal mechanism keeps invariant across environments, while the spurious correlation varies, and aim to learn representations that are predictive across multiple environments, e.g., data from different hospitals or acquisition settings~\cite{zare2022removal}. Invariant learning also extends to the scenario where environment labels are unknown and prior knowledge of spurious correlation is used to split the training data into groups, known as group invariant learning~\cite{chen2022does, yang2024identifying}.

\subsection{Model-centric techniques}
Model-centric approaches seek to mitigate shortcut learning by modifying the training objective, model architecture, or inference procedure. 
In-processing methods during training include explanation-based regularization, which leverages model explanations or attribution maps as proxies for identifying shortcut features and penalizes reliance on them~\cite{ross2017right, schramowski2020making, shao2021right}. Group robustness approaches, most notably Group Distributionally Robust Optimization (Group DRO), use group annotations to minimize worst-group performance and reduce reliance on shortcuts that fail for underrepresented subpopulations~\cite{sagawa2019distributionally, zhou2021examining}. Knowledge distillation-based methods, originally proposed for model compression~\cite{hinton2015distilling}, have also been adapted for bias and shortcut mitigation by training teacher models on curated or balanced data and transferring this knowledge to student models~\cite{boland2024all, bassi2024explanation, kenfack2024adaptive, tian2024distilling}, including recent extensions that distill intermediate representations to prevent shortcut learning~\cite{cha2022domain, boland2025preventing}.

Post-processing methods during inference-time aim to mitigate shortcuts without retraining the model, which is particularly useful when access to the full training data is limited. These include concept bottleneck models~\cite{koh2020concept}, where interventions on intermediate concept representations can suppress shortcut reliance~\cite{stammer2024neural}, pruning methods that remove neurons associated with biased predictions~\cite{wu2022fairprune}, fine-tuning strategies that mask or reweight shortcut-related parameters~\cite{asgari2022masktune, xue2025bmft}, and post-hoc techniques to disentangle content and style in learned representations from pre-trained vision models to prevent style-related spurious correlations~\cite{ngweta2023simple}.

\subsection{Feature disentanglement}
Feature disentanglement constitutes a class of model-centric in-processing techniques that explicitly aim to separate task-relevant information from confounding or spurious factors in the learned representation. Adversarial training approaches enforce invariance by training a discriminator to predict confounders from latent representations while the encoder attempts to remove this information~\cite{ganin2015unsupervised}, with successful applications in domain adaptation in medical imaging~\cite{kamnitsas2017unsupervised, he2021adversarial}. 

Alternatively, latent space splitting decomposes representations explicitly into multiple subspaces corresponding to different factors of variation. This has been studied using generative models such as variational autoencoders (VAEs)~\cite{yang2022chroma} and generative adversarial networks (GANs)~\cite{xie2020mi, mueller2024disent}, as well as encoder-based architectures in classification settings~\cite{fay2023avoid, mueller2025bench, wang2024navigate}. Latent space splitting is typically combined with dependence estimators to encourage statistical independence between subspaces, including mutual information minimization (MI)~\cite{xie2020mi, yang2022chroma, fay2023avoid}, distance correlation (dCor)~\cite{sz2007dCor, zhen2022versatile, mueller2024disent}, and non-parametric kernel-based measures such as maximum mean discrepancy (MMD)~\cite{gretton2012kernel}, as used for example in the Variational Fair Autoencoder~\cite{louizos2015variational}.

\subsection{Scope of this work}
In this paper, we focus on representation-level methods, as shortcut mitigation ultimately depends on how spurious and task-relevant factors are encoded in the learned feature space. In particular, we center our analysis on feature disentanglement approaches, for which benchmarking is especially informative since these methods share a common latent space formulation while differing in how dependence between features is reduced. This enables controlled and meaningful comparisons within a single methodological subfield. Within this scope, we benchmark a data-centric rebalancing strategy via oversampling, adversarial learning, and latent space splitting approaches based on explicit dependence minimization using distance correlation, mutual information, and maximum mean discrepancy against a standard empirical risk minimization (ERM) baseline. Finally, we investigate the interaction between data-centric and representation-level techniques by combining rebalancing with each method to assess their individual and combined effectiveness.

\section{Methods}
\subsection{Problem setting}
We considered a multi-task classification setting with two binary prediction tasks: a primary task \(y_1\) and a spuriously correlated auxiliary task \(y_2\), which acted as a confounder. Both tasks were predicted from a shared image input \(x\) (Fig.\,\ref{fig:overview-figure}\,\textbf{c}), and shortcut learning occurred when the model exploited correlations between \(y_1\) and \(y_2\) instead of task-relevant image features. To mitigate this effect, we adopted a latent space splitting approach based on feature disentanglement. An encoder \(f_\theta\) mapped the input image to a latent representation that was explicitly divided into task-specific subspaces
\begin{equation}
    f_\theta(x) = z = (z_1, z_2),
\end{equation}
where \(z_1\) encoded information relevant to the primary task \(y_1\), and \(z_2\) encoded features related to the confounding task \(y_2\). Each task was predicted from its corresponding latent subspace using a linear classification head,
\begin{equation}
    \hat{y}_i = C_{\psi_i}(z_i), \quad i \in \{1,2\}.
\end{equation}
The classification objective was defined as the average cross-entropy loss over both tasks
\begin{align}
    (\theta^*, \psi^*) &= \argmin_{\theta, \psi} \mathcal{L}_{\text{cls}}(\theta, \psi),\label{eq:cls-loss} \\
    \mathcal{L}_{\text{cls}}(\theta, \psi) &= \frac{1}{2} \left( \mathcal{L}_{\text{CE}}(\theta, \psi_1) +\mathcal{L}_{\text{CE}}(\theta, \psi_2) \right),\\
    \mathcal{L}_{\text{CE}}(\theta, \psi_i) &= \mathbb{E}_{(x,y_i)}\big[-\, y_i^\top \log C_{\psi_i}(z_i)\big].
\end{align}
To explicitly enforce disentanglement between subspaces, we penalized statistical dependence between the latent subspaces using a dependence measure \(D(\cdot,\cdot)\). The resulting optimization problem was
\begin{align}
    \mathcal{L}_{\text{dep}}(\theta) &= \mathbb{E}_{x} \big[ D(z_1, z_2)\big], \\
    (\theta^*, \psi^*) &= \argmin_{\theta, \psi} \mathcal{L}_{\text{cls}}(\theta, \psi) + \lambda \mathcal{L}_{\text{dep}}(\theta),
\end{align}
where \(\lambda\) controls the trade-off between task performance and disentanglement strength.

\subsection{Baseline strategies}
As a baseline, we employed standard empirical risk minimization (ERM) that optimized only the classification loss without any explicit mitigation of shortcut learning (Eq.~\ref{eq:cls-loss}). In addition, we considered a data-centric rebalancing strategy that addressed spurious correlations at the dataset level. Specifically, we oversampled with replacement underrepresented samples in the contingency table defined by the joint distribution of the primary task \(y_1\) and the confounder \(y_2\). This reduced correlation between tasks in the training data and serves both as a standalone mitigation technique and as a complementary component combined with model-centric disentanglement methods.

\subsection{Adversarial learning}
Adversarial learning mitigates shortcut learning by enforcing invariance of the learned representation with respect to confounding factors through a minimax optimization objective. We refer to the adversarial classifier approach of~\cite{ganin2015unsupervised} as AdvCl. Instead of explicitly minimizing a statistical dependence measure, AdvCl suppressed confounder-related information implicitly through competing objectives.
Following~\cite{ganin2015unsupervised}, two classifiers operated on a shared latent representation \(z\): one predicting the target task \(y_1\) and one predicting the confounder \(y_2\). The training objective was
\begin{align}
    \mathcal{L}(\theta, \psi_1, \psi_2) &= \mathcal{L}_{\text{CE}}(\theta, \psi_1) - \lambda \mathcal{L}_{\text{CE}}(\theta, \psi_2), \\
    \mathcal{L}_{\text{CE}}(\theta, \psi_i) &= \mathbb{E}_{(x,y_i)}\big[-\, y_i^\top \log C_{\psi_i}(z)\big],
\end{align}
and was optimized as an adversarial game:
\begin{align}
    (\theta^*, \psi_1^*) &= \argmin_{\theta, \psi_1} \mathcal{L}(\theta, \psi_1, \psi_2^*), \\
    \psi_2^* &= \argmax_{\psi_2} \mathcal{L}(\theta^*, \psi_1^*, \psi_2).
\end{align}
For fixed \(\theta\), the inner maximization makes \(C_{\psi_2}\) the optimal predictor of \(y_2\) from \(z\). The outer minimization then updates \(\theta\) to increase this optimal adversarial loss, which is only possible if \(y_2\) becomes unpredictable from \(z\). Thus, the latent representation removes \(y_2\)-informative structure while preserving information relevant for predicting \(y_1\). In practice, this minimax problem was efficiently optimized using a gradient reversal layer (GRL), which acted as the identity during the forward pass and reversed the gradient sign during backpropagation~\cite{ganin2015unsupervised}. This formulation encouraged representations that were predictive for the target task while being invariant to the confounder.

\subsection{Feature disentanglement approaches}
Following adversarial learning, which enforced invariance implicitly via a minimax objective, we next considered feature disentanglement approaches that explicitly minimized statistical dependence between latent representations. In the context of shortcut learning, these methods reduce unintended coupling between task-relevant and confounder-related features by constraining the relationship between latent subspaces \(z_1\) and \(z_2\). By penalizing statistical dependence during training, they encourage complementary, minimally redundant representations, limiting reliance on spurious correlations.

\subsubsection{Distance correlation}
Distance correlation (dCor)~\cite{sz2007dCor}, proposed by~\cite{mueller2024disent, zhen2022versatile} for feature disentanglement, measures both linear and nonlinear dependence between random vectors of arbitrary dimension and is zero if and only if the variables are statistically independent. Given batch samples \(z_1 \in \mathbb{R}^{N \times d_1}\) and \(z_2 \in \mathbb{R}^{N \times d_2}\), empirical distance correlation is defined as
\begin{align}
    D_{\text{dCor}}(z_1, z_2) &= \frac{\text{dCov}(z_1, z_2)}{\sqrt{\text{dCov}(z_1, z_1)\,\text{dCov}(z_2, z_2)}}, \\
    \text{dCov}(z_1, z_2) &= \sqrt{\sum_{i=1}^N \sum_{j=1}^N \frac{A_{ij} B_{ij}}{N^2}},
\end{align}
where \(A\) and \(B\) were centered pairwise Euclidean distance matrices computed from \(z_1\) and \(z_2\), respectively.

\subsubsection{Mutual information neural estimation}
Mutual information (MI) quantifies the statistical dependence between two random variables,
\begin{equation}
    \text{MI}(z_1, z_2) = D_{\text{KL}}(P_{z_1, z_2} \,\|\, P_{z_1} P_{z_2}),
\end{equation}
and equals zero if and only if the variables are independent. The mutual information neural estimator (MINE)~\cite{belghazi2018mutual} provides a lower bound on MI by optimizing
\begin{equation}
    D_{\text{KL}}(\mathbb{P} \,\|\, \mathbb{Q}) \geq \sup_{T \in \mathcal{F}} \mathbb{E}_{\mathbb{P}}[T] - \log \mathbb{E}_{\mathbb{Q}}[e^T],
\end{equation}
where \(T_\theta : Z_1 \times Z_2 \rightarrow \mathbb{R}\) is a neural network parameterized by \(\theta\). The resulting dependence measure is
\begin{equation}
    D_{\text{MI}}(z_1, z_2) = \sup_{\theta} \mathbb{E}_{P_{z_1, z_2}}[T_\theta] - \log \mathbb{E}_{P_{z_1} P_{z_2}}[e^{T_\theta}],
\end{equation}
which was initial proposed as a feature disentanglement measure for medical imaging in~\cite{fay2023avoid}.

\subsubsection{Maximum mean discrepancy}
Maximum mean discrepancy (MMD)~\cite{gretton2012kernel} is a non-parametric, kernel-based measure of discrepancy between probability distributions and is zero if and only if the distributions are identical in a reproducing kernel Hilbert space. We used MMD to encourage statistical independence between latent subspaces by minimizing the discrepancy between their empirical distributions. Given samples \(z_1 \in \mathbb{R}^{N \times d_1}\) and \(z_2 \in \mathbb{R}^{N \times d_2}\), the unbiased quadratic-time estimator is
\begin{align}
	D_{\text{MMD}}(z_1, z_2) &= \frac{1}{N(N-1)} \sum_{i \neq j} k(z_1^{(i)}, z_1^{(j)}) + \frac{1}{N(N-1)} \sum_{i \neq j} k(z_2^{(i)}, z_2^{(j)}) - \frac{2}{N^2} \sum_{i=1}^N \sum_{j=1}^N k(z_1^{(i)}, z_2^{(j)}),
\end{align}
where \(k(\cdot,\cdot)\) is a kernel function. We employed a mixture of Gaussian radial basis function kernels with multiple bandwidths (from \(2^{-3}\) to \(2^{3}\)) to capture discrepancies at different scales.

\subsection{Experimental design}
We evaluated shortcut mitigation methods in a multi-task classification setup with a primary task \(y_1\) and a confounder \(y_2\). Both tasks were predicted from the same images, and shortcut learning occurs when models exploit correlations between \(y_1\) and \(y_2\) instead of task-relevant features. Our experiments systematically assessed the ability of different methods to mitigate such shortcuts across controlled and real-world datasets.

\subsubsection{Datasets}
We analyzed three publicly available datasets, including one controlled toy dataset and two medical imaging datasets from radiology and ophthalmology with natural and synthetic confounders, respectively (Table~\ref{tab:experiments}).

\textbf{Morpho-MNIST}\footnote{\url{https://github.com/dccastro/Morpho-MNIST}}~\cite{castro2019morpho_mnist}: We used \(28\times28\) grayscale digits from the ``global'' dataset, selecting thin and thick digits (39,980 training and 6,693 test samples). The primary task \(y_1\) was a binary digit classification (0–4 versus 5–9) and the confounder \(y_2\) was writing style (thin versus thick).  

\textbf{CheXpert}\footnote{\url{https://stanfordmlgroup.github.io/competitions/chexpert/}}~\cite{irvin2019chexpert}: We used frontal chest radiographs (39,979 patients, 100,014 training images; 827 patients, 2,183 test images), resized and cropped to \(320 \times 320\). The primary task was detecting pleural effusion (\(y_1\)), and the confounder was patient sex (\(y_2\)).

\textbf{OCT}\footnote{\url{https://data.mendeley.com/datasets/rscbjbr9sj/2}}~\cite{kermany2018oct}: Optical coherence tomography (OCT) images were cropped to a central square and resampled to \(256 \times 256\) resolution (3,949 patients, 59,756 training B-Scans; 225 patients, 500 test B-Scans). The primary task was detecting drusen (\(y_1\)), and we synthetically applied a confounder (\(y_2\)) to simulate image acquisition noise. The synthetic confounder was implemented by applying a radial (donut-shaped) notch filter in the frequency domain. The filter suppressed a narrow band of spatial frequencies centered at 55\% of the maximum radial frequency using a smooth Gaussian attenuation. The notch strength was set to 0.9 to strongly attenuate the target frequencies while minimizing ringing artifacts in the spatial domain.

\begin{table}[htbp]
\centering
\caption{Summary of the datasets and tasks.}
\label{tab:experiments}
\setlength{\tabcolsep}{4pt}
\begin{tabularx}{\textwidth}{l X X}
    \toprule
    Dataset & Primary task \(y_1\) & Confounder \(y_2\) \\
    \midrule
    Morpho-MNIST & digits (0--4 / 5--9) & style (thin / thick) \\
    CheXpert & pleural effusion (no / yes) & sex (female / male) \\
    OCT & drusen (no / yes) & radial notch (no / yes) \\
    \bottomrule
\end{tabularx}
\end{table}

\subsubsection{Data distributions}
\begin{figure}[ht]
\centering
\begin{minipage}[t]{0.48\textwidth}
    \centering
    \includegraphics[width=\textwidth]{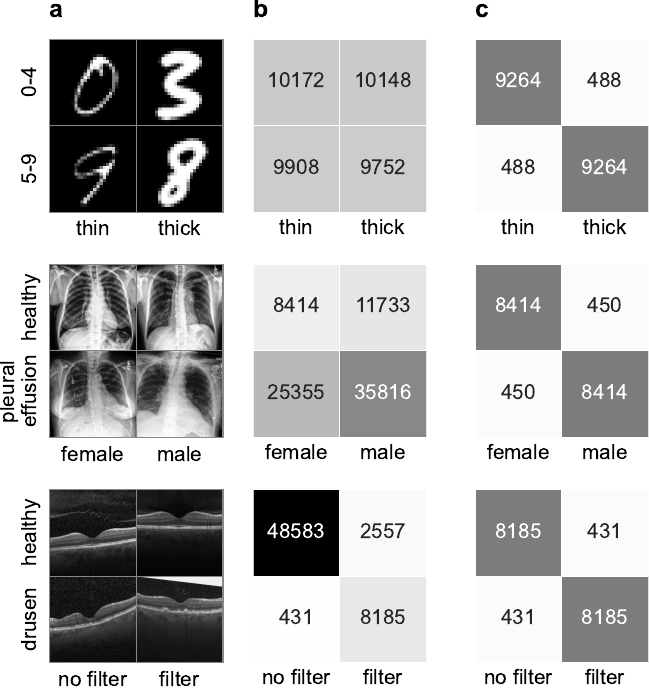}
    \caption{Overview of label distributions in Morpho-MNIST, CheXpert, and OCT. \textbf{a} shows example images sampled for each label combination, \textbf{b} shows contingency tables of the original training data, and \textbf{c} shows contingency tables of the subsampled training data actually used. In the final training data (\textbf{c}), strong correlations were induced between the primary task and confounder for all datasets, while maintaining balanced distributions within each classification task.}
    \label{fig:data_distributions}
\end{minipage}
\hfill
\begin{minipage}[t]{0.48\textwidth}
    \centering
    \includegraphics[width=\textwidth]{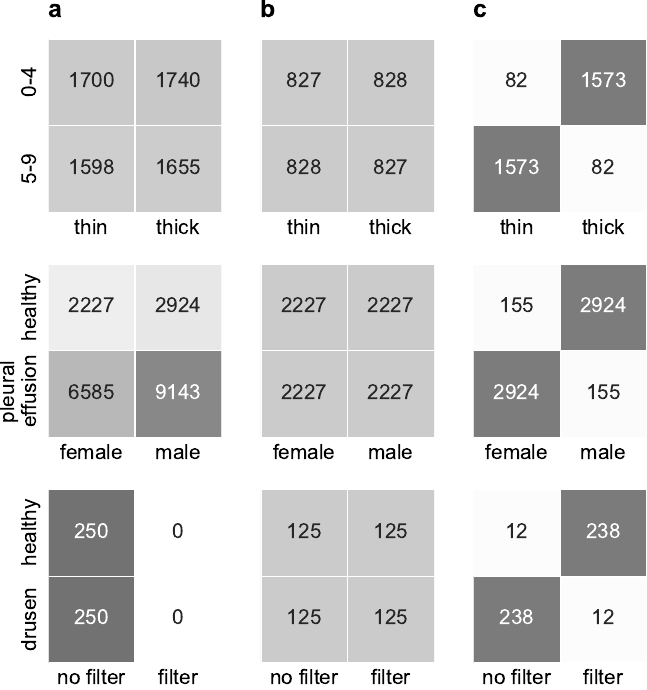}
    \caption{Test data distributions for evaluating shortcut mitigation, showing original (\textbf{a}), balanced (\textbf{b}), and inverted (\textbf{c}) correlations between the primary task and confounder.}
    \label{fig:test_data_distributions}
\end{minipage}
\end{figure}

For each dataset, we created a strongly correlated training distributions by subsampling the original data distributions (Fig.~\ref{fig:data_distributions}). The marginal frequencies of both binary labels remain approximately balanced (50/50), but the labels were internally correlated. To simulate strong shortcut learning, 95\% of the samples were selected from the main diagonal of the co-occurrence matrix (i.e., matching \(y_1\) and \(y_2\) categories), with the remaining 5\% sampled from off-diagonal entries (Fig.~\ref{fig:data_distributions}\textbf{c}). For example, in Morpho-MNIST, 95\% of training images were small-thin or high-thick digits, with the remaining 5\% from other label combinations. For the OCT dataset, as a special case where the original dataset is not confounded, we first created a synthetically confounded dataset in which 95\% of the data lies on the main diagonal of the co-occurrence matrix (Fig.~\ref{fig:data_distributions}\textbf{b}, last row), and then subsampled it in a second step to obtain internally balanced tasks (Fig.~\ref{fig:data_distributions}\textbf{c}, last row).

\subsubsection{Evaluation protocol}\label{sec:evaluation_protocol}
All methods were trained using 5-fold cross-validation. For the medical datasets, the training, and validation splits were separated by patient identifiers to avoid data leakage. The best model in each fold was selected based on the validation loss. Evaluation was conducted on three types of test distributions:

\begin{itemize}
    \item \textbf{Original}: The standard test set provided by the dataset. For the OCT dataset, none of the original images have the synthetic confounder applied.
    \item \textbf{Balanced}: A subsampled test set with no correlation between \(y_1\) and \(y_2\). 
    \item \textbf{Inverted}: A subsampled test set in which the correlation between \(y_1\) and \(y_2\) is reversed relative to training, such that 95\% of samples correspond to label-confounder combinations that were rare during training.

\end{itemize}
We reported the actual numbers of images of the original, balanced, and inverted test distributions (Fig.~\ref{fig:test_data_distributions}~\textbf{a}, \textbf{b}, and \textbf{c}, respectively).
Methods that successfully mitigate shortcut learning are expected to maintain consistent performance across these test distributions.

\subsubsection{Training details}
All models were implemented in PyTorch Lightning~\cite{falcon2019lightning} and trained on a single NVIDIA TITAN Xp GPU. Training hyperparameters were selected per dataset and method; exact configurations are provided in the accompanying code repository, as they varied across experiments. Across all experiments, the encoder output a 4-dimensional latent representation, which was split into two 2-dimensional subspaces corresponding to the primary task \(y_1\) and the confounder \(y_2\). Task predictions were obtained via linear classification heads that mapped from the respective 2-dimensional latent subspaces to the binary task outputs.

For Morpho-MNIST, we employed the same encoder architecture as in~\cite{mueller2025bench,fay2023avoid}, consisting of three convolutional layers followed by a linear projection to the 4-dimensional latent space. Models were trained with a batch size of 900 using the Adam optimizer (learning rate \(10^{-3}\), weight decay \(10^{-5}\)). For CheXpert and OCT, we used a ResNet18 backbone (without pre-trained weights) with the final fully connected layer replaced by a linear projection to the 4-dimensional latent space. Training was performed using AdamW with a learning rate of \(10^{-3}\) and a weight decay of \(10^{-2}\), using batch sizes of 50 for CheXpert and 256 for OCT.

The training schedule for MINE differed from that of other feature disentanglement methods. Specifically, MINE was optimized alternately: in each iteration, after the encoder was updated, the MI estimator network was updated for \(N_B-1\) additional batches, resulting in \(N_B\)-times more effective training epochs compared to other methods. Due to different convergence properties of the compared methods, the total number of training epochs varied between them. Therefore, the number of epochs as well as further hyperparameters are provided with the code\footnote{\url{https://github.com/berenslab/medical-shortcut-mitigation}}.

\section{Results}

\subsection{Shortcut mitigation methods improve classification performance of the primary diagnostic task}

To benchmark the effectiveness of different approaches for mitigating shortcut learning, we first evaluated classification performance on the primary diagnostic task across three datasets exhibiting varying degrees of spurious correlations. Performance was measured using the area under the receiver operating characteristic curve (AUROC). We report AUROC for the primary task \(y_1\), inferred from the corresponding latent subspace \(z_1\) (Table~\ref{tab:classification-performance}). As a threshold-independent metric, AUROC enables consistent comparison across classifiers with different decision boundaries. All results are reported as percentages, along with the standard deviation over 5-fold cross-validation. We evaluated generalization under three test distributions---\emph{Original}, \emph{Balanced}, and \emph{Inverted}---which differed in the correlation structure between the primary label \(y_1\) and the confounder \(y_2\) (see Section~\ref{sec:evaluation_protocol}, Fig.~\ref{fig:test_data_distributions}). We expected methods that successfully mitigate shortcut learning to maintain strong and stable performance across all three distributions.

\begin{table}[htbp]
\centering
\caption{Classification performance of the main task (\(z_1 \rightarrow y_1\)) measured as AUROC over different datasets and distributions.}
\label{tab:classification-performance}
\setlength{\tabcolsep}{4pt}      
\renewcommand{\arraystretch}{1.1} 
\begin{tabularx}{\textwidth}{l *{9}{>{\centering\arraybackslash}X}} 
\toprule
 & \multicolumn{3}{c}{Morpho-MNIST (digits)} 
 & \multicolumn{3}{c}{CheXpert (pleural effusion)} 
 & \multicolumn{3}{c}{OCT (drusen)} \\
\cmidrule(lr){2-4} \cmidrule(lr){5-7} \cmidrule(lr){8-10}
Method & Original & Balanced & Inverted 
       & Original & Balanced & Inverted 
       & Original & Balanced & Inverted \\
\midrule
Baseline (ERM)  & \(96 \pm 1\) & \(95 \pm 1\) & \(88 \pm 3\) 
                & \(79 \pm 1\) & \(79 \pm 1\) & \(46 \pm 4\) 
                & \(99 \pm 1\) & \(92 \pm 5\) & \(74 \pm 15\)  \\
Rebalancing & \(98 \pm 1\) & \(97 \pm 1\) & \(97 \pm 0\) 
                    & \(88 \pm 2\) & \(88 \pm 2\) & \(84 \pm 5\) 
                    & \(99 \pm 1\) & \(96 \pm 3\) & \(90 \pm 9\)  \\
AdvCl       & \(98 \pm 0\) & \(97 \pm 1\) & \(96 \pm 1\)
            & \(83 \pm 1\) & \(83 \pm 1\) & \(64 \pm 4\) 
            & \(99 \pm 1\) & \(97 \pm 2\) & \(93 \pm 5\)  \\
AdvCl+Rebal & \(98 \pm 2\) & \(97 \pm 2\) & \(96 \pm 1\)
            & \(85 \pm 2\) & \(85 \pm 1\) & \(72 \pm 9\) 
            & \(99 \pm 1\) & \(97 \pm 1\) & \(92 \pm 4\)  \\
dCor        & \(97 \pm 0\) & \(96 \pm 1\) & \(92 \pm 1\)
            & \(79 \pm 3\) & \(79 \pm 3\) & \(49 \pm 9\) 
            & \(99 \pm 2\) & \(95 \pm 2\) & \(88 \pm 5\)  \\
dCor+Rebal  & \(99 \pm 0\) & \(98 \pm 0\) & \(98 \pm 0\)
            & \(88 \pm 1\) & \(88 \pm 1\) & \(85 \pm 4\) 
            & \(99 \pm 2\) & \(99 \pm 0\) & \(98 \pm 0\) \\
MINE        & \(97 \pm 2\) & \(96 \pm 2\) & \(98 \pm 1\)
            & \(88 \pm 1\) & \(88 \pm 1\) & \(85 \pm 3\) 
            & \(98 \pm 1\) & \(97 \pm 1\) & \(97 \pm 1\)  \\
MINE+Rebal  & \(97 \pm 1\) & \(96 \pm 1\) & \(97 \pm 1\)
            & \(86 \pm 1\) & \(86 \pm 1\) & \(84 \pm 3\) 
            & \(99 \pm 1\) & \(98 \pm 1\) & \(97 \pm 2\)  \\
MMD         & \(97 \pm 0\) & \(96 \pm 0\) & \(91 \pm 1\)
            & \(74 \pm 2\) & \(74 \pm 2\) & \(49 \pm 4\) 
            & \(99 \pm 0\) & \(94 \pm 1\) & \(82 \pm 3\)  \\
MMD+Rebal   & \(98 \pm 0\) & \(97 \pm 0\) & \(97 \pm 1\)
            & \(82 \pm 1\) & \(82 \pm 1\) & \(77 \pm 3\) 
            & \(99 \pm 1\) & \(97 \pm 1\) & \(93 \pm 4\)  \\
\bottomrule
\end{tabularx}
\end{table}

\paragraph{Artificial dataset (Morpho-MNIST)}
For the Morpho-MNIST dataset, we considered binary digit classification confounded by writing style, operationalized via line thickness. On the \emph{original} test distribution (Fig.~\ref{fig:test_data_distributions}\textbf{a}, first row; Table~\ref{tab:classification-performance}, first column), AUROC ranges from approximately 96\% for the baseline to 99\% for dCor+Rebal. All proposed methods improved over the baseline by 1--3 percentage points, with dCor+Rebal achieving the best performance.
On the \emph{balanced} test distribution (Fig.~\ref{fig:test_data_distributions}\textbf{b}, first row; Table~\ref{tab:classification-performance}, first column), where the correlation between digit and writing style was removed, dCor+Rebal again performed best, improving AUROC by 4\% over the baseline. The \emph{inverted} test distribution (Fig.~\ref{fig:test_data_distributions}\textbf{c}, first row; Table~\ref{tab:classification-performance}, first column) is the most informative setting for assessing shortcut learning, as the correlation between digit and writing style is reversed relative to training. Therefore, a classifier relying (solely) on the shortcut will suffer severe drops in performance. Here, performance gains for shortcut mitigation methods were most pronounced over the baseline: dCor+Rebal and MINE achieved the strongest improvements, each outperforming the baseline by approximately 10\%. All other methods also yielded significant gains of between 3\% and 9\%. Although combining feature disentanglement methods with Rebalancing was beneficial in some cases, only the combination with dCor led to larger improvements than Rebalancing alone.

\paragraph{Radiology task (CheXpert)}
For CheXpert, we evaluated pleural effusion classification with patient sex as a confounder. On the \emph{original} test distribution (Fig.~\ref{fig:test_data_distributions}\textbf{a}, second row; Table~\ref{tab:classification-performance}, second column), improvements over the baseline were more pronounced than for Morpho-MNIST, reaching up to +9\% for Rebalancing, dCor+Rebal, and MINE. In contrast, MMD performed worse than the baseline. Combining several methods with Rebalancing generally improved performance, but did not improve upon Rebalancing alone. Performance on the \emph{balanced} test distribution (Fig.~\ref{fig:test_data_distributions}\textbf{b}, second row; Table~\ref{tab:classification-performance}, second column) closely mirrored that of the original distribution. This suggests that, for CheXpert, the models were less sensitive to the confounder under the original test setting, such that removing the correlation between pleural effusion and patient sex had a smaller effect on classification performance. Notably, the confounder in this dataset was naturally occurring and exhibited the weakest predictive power among the three datasets considered (Table~\ref{tab:disent-performance}). In contrast, when the correlation between pleural effusion and patient sex was not merely removed but actively inverted, the effect of shortcut reliance became clearly visible. On the \emph{inverted} test distribution (Fig.~\ref{fig:test_data_distributions}\textbf{c}, second row; Table~\ref{tab:classification-performance}, second column), baseline performance collapsed to 46\%, accompanied by overall higher variance than for the other distributions. Rebalancing alone achieved a 38\% improvement in AUROC over baseline performance, closely matching the performance of dCor+Rebal and MINE. In contrast, AdvCl and its combination with Rebalancing performed significantly worse, achieving only an improvement of 18\% and 26\%, respectively. Notably, combining dCor with Rebalancing resulted in a 36\% improvement over dCor alone, highlighting the strong complementary effect of data-centric and model-centric shortcut mitigation strategies.

\paragraph{Ophthalmology task (OCT)}
For the OCT dataset, we considered drusen classification confounded by a synthetic radial notch filter. On the \emph{original} test distribution (Fig.~\ref{fig:test_data_distributions}\textbf{a}, third row; Table~\ref{tab:classification-performance}, third column), where no confounder was applied, all methods achieved consistently high performance of approximately 99\% AUROC, indicating that the task is easily solvable in the absence of confounding. When evaluating the \emph{balanced} test distribution (Fig.~\ref{fig:test_data_distributions}\textbf{b}, third row; Table~\ref{tab:classification-performance}, third column), in which the radial notch filter was applied to 50\% of the images, baseline performance dropped to 92\%. All shortcut mitigation methods recovered this loss to varying degrees, achieving AUROC improvements between 2\% and 5\%. dCor+Rebal again performed best with +5\%, effectively matching the performance observed in the unconfounded setting, followed closely by MINE+Rebal with +4\%. The impact of shortcut mitigation was most pronounced on the \emph{inverted} test distribution (Fig.~\ref{fig:test_data_distributions}\textbf{c}, third row; Table~\ref{tab:classification-performance}, third column), where the confounder--label correlation was reversed. Baseline performance dropped to 74\% with a high standard deviation of 15\%. This decline was most effectively addressed by dCor+Rebal, which improved performance by 24\% with low variance across folds. MINE and MINE+Rebal achieved comparable gains of 23\%. In contrast, the next best methods---AdvCl and MMD+Rebal---yielded improvements of approximately 19\% over the baseline, highlighting a more pronounced performance gap in this setting.

\paragraph{Summary}
Overall, these results show that when multi-task classifiers are trained under strong task correlations---where 95\% of training samples lie on the main diagonal of the task co-occurrence matrix---the baseline model is highly vulnerable to distribution shifts, exhibiting substantial performance drops across altered test distributions. In contrast, both data-centric rebalancing and model-centric feature disentanglement methods, including adversarial learning (AdvCl), consistently improved classification performance over the baseline, as measured by AUROC, across all evaluated test settings. These gains are most pronounced on the inverted test distribution, where the correlation between tasks is reversed relative to training. Notably, combining data-centric and model-centric interventions yields further improvements beyond either strategy in isolation. Across all three datasets and test distributions, the most consistent and robust performance was achieved by combining Rebalancing with distance correlation (dCor)--based latent space disentanglement, highlighting the complementary benefits of jointly addressing shortcut learning at both the data and representation levels.

\subsection{Beyond classification metrics: latent space analysis reveals differences in feature disentanglement}

Next, we evaluated how well the different shortcut mitigation methods disentangled the two spaces \(z_1\) and \(z_2\) across all datasets and methods, both quantitatively and qualitatively. Quantitatively, we used a \(k\)-nearest neighbor (\(k\)NN) classifier with \(k=30\) to assess how well each latent subspace encodes each label. In the subspace-label confusion matrices evaluated on the \emph{balanced} test distribution, off-diagonal values near 50\% indicate optimal disentanglement with minimal confounder leakage into the task subspace (Tab.~\ref{tab:disent-performance}). For the AdvCl, which uses a single latent space made invariant to the confounder, the confusion matrix contains only two entries, reflecting the single shared latent space.

\begin{table}[htbp]
\small
\centering
\caption{Disentanglement performance in subspace-label confusion matrices. Performance is measured as \(k\)NN classification accuracy (\(k=30\)) across all subspace-label combinations, datasets, and methods. Darker shades indicate higher accuracy; off-diagonal values near 50\% reflect optimal disentanglement with minimal confounder leakage.}
\label{tab:disent-performance}
\setlength{\tabcolsep}{5pt}    
\renewcommand{\arraystretch}{1.1}
\begin{tabularx}{\textwidth}{l *{10}{>{\centering\arraybackslash}X}}
\toprule
Dataset & Baseline & Rebal & AdvCl & \shortstack{AdvCl+\\Rebal} 
        & dCor & \shortstack{dCor+\\Rebal} & MINE & \shortstack{MINE+\\Rebal} 
        & MMD & \shortstack{MMD+\\Rebal} \\
\midrule
Morpho-MNIST &
    \cvconfplot{87}{52}{80}{100} &
    \cvconfplot{86}{50}{84}{100} &
    \cvconfplotone{91}{59} &
    \cvconfplotone{91}{55} &
    \cvconfplot{90}{52}{64}{99} &
    \cvconfplot{93}{50}{52}{100} &
    \cvconfplot{90}{50}{48}{100} &
    \cvconfplot{90}{50}{52}{100} &
    \cvconfplot{89}{54}{72}{99} &
    \cvconfplot{84}{58}{98}{100} \\
CheXpert &
    \cvconfplot{68}{63}{85}{89} &
    \cvconfplot{76}{57}{67}{90} &
    \cvconfplotone{74}{69} &
    \cvconfplotone{77}{61} &
    \cvconfplot{71}{56}{76}{90} &
    \cvconfplot{80}{54}{54}{86} &
    \cvconfplot{80}{50}{52}{90} &
    \cvconfplot{80}{51}{50}{90} &
    \cvconfplot{71}{61}{82}{93} &
    \cvconfplot{73}{60}{60}{77} \\
OCT &
    \cvconfplot{84}{61}{84}{95} &
    \cvconfplot{90}{54}{80}{97} &
    \cvconfplotone{91}{76} &
    \cvconfplotone{92}{65} &
    \cvconfplot{88}{52}{77}{96} &
    \cvconfplot{94}{51}{55}{95} &
    \cvconfplot{91}{50}{53}{95} &
    \cvconfplot{94}{51}{54}{96} &
    \cvconfplot{87}{76}{88}{96} &
    \cvconfplot{85}{80}{96}{96} \\
\bottomrule
\end{tabularx}
\end{table}

\begin{figure*}[htbp]
    \centering
    \includegraphics[width=\linewidth]{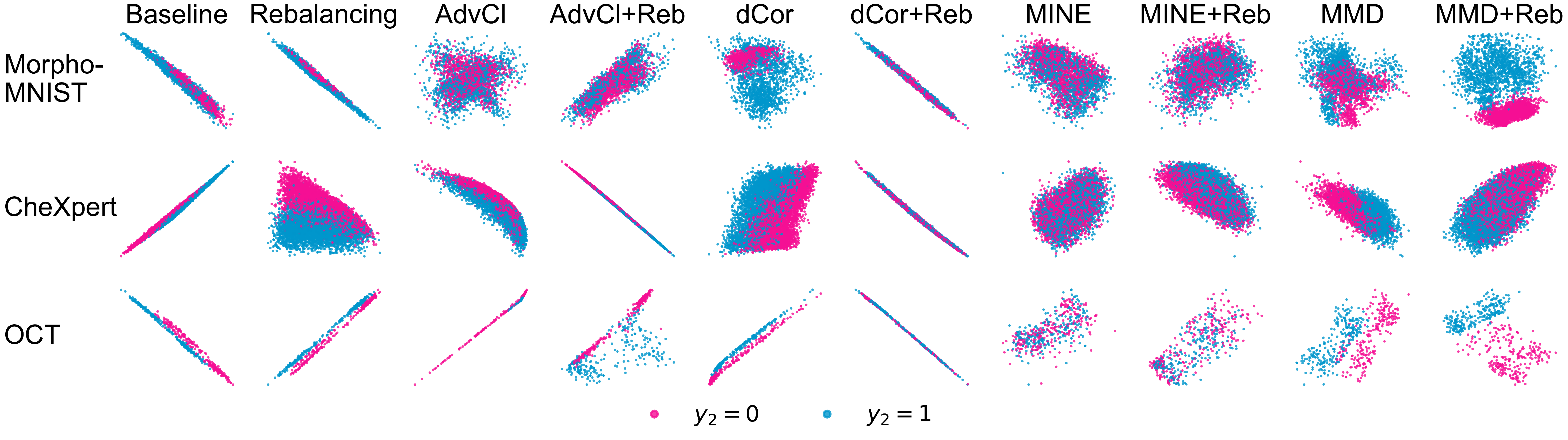}
    \caption{Qualitative scatter plots showing the two-dimensional subspace \(z_1\) of the first fold, highlighted by the binary labels \(y_2\). Feature disentanglement was successful if there is no structure visible with respect to \(y_2\).}
    \label{fig:disent-performance}
\end{figure*}

\paragraph{Artificial dataset (Morpho-MNIST)}
For Morpho-MNIST, the baseline and the Rebalancing method exhibited poor disentanglement, with the main task subspace \(z_1\) still encoding substantial confounder information, indicated by high off-diagonal accuracy. In contrast, combining dCor with Rebalancing showed the most clear benefits, with off-diagonal values closer to 50\%, which also showed in strong AUROC performance (Table~\ref{tab:classification-performance}). MMD, however, performed poorly in this metric, and MMD+Rebal was even worse than the baseline. 

\paragraph{Radiology task (CheXpert)}
For CheXpert, \(k\)NN accuracy on the main diagonals was lower than in MNIST, indicating that both task and confounder prediction was more challenging. However, the baseline showed evidence for a strong presence of the confounder in the task space, with similar accuracies across rows. Rebalancing provided some improvement, despite not explicitly enforcing disentanglement through a dependence-minimizing regularization. However, stronger disentanglement could be achieved by methods that explicitly target latent independence, with dCor+Rebal, MINE, and MINE+Rebal showing the most pronounced improvements.

\paragraph{Ophthalmology task (OCT)}
For OCT, diagonal accuracies were again comparable to MNIST, with the baseline confusion matrix showing substantial encoding of confounder information in the task subspace. Rebalancing alone did not substantially improve disentanglement. As for the other two datasets, the best performance was achieved by dCor+Rebal, MINE, and MINE+Rebal, while MMD and MMD+Rebal failed to produce convincing disentanglement results.

\paragraph{Summary}
Across all datasets and test distributions, prediction of the confounder \(y_2\) remained consistently high, as evidenced by the corresponding main-diagonal entries in the subspace-label confusion matrices. This indicates that the confounding signal was present and readily learnable in all settings. Overall, similar AUROC values across methods in Table~\ref{tab:classification-performance} can mask substantial differences in representation quality. Rebalancing often matched the AUROC of model-centric approaches but did not consistently yield effective subspace disentanglement (Table~\ref{tab:disent-performance}). In contrast, strong disentanglement---indicated by off-diagonal accuracies near 50\% and high \(k\)NN accuracy on the main diagonal---also coincided with high AUROC across data distributions. Moreover, methods that explicitly minimized statistical dependencies between latent subspaces achieved better separation of task and confounder information. Combining model-centric disentanglement with data-centric interventions further improved performance for dCor and AdvCl, while this combination proved disadvantageous for MMD. Across all datasets, the strongest disentanglement performance was achieved by dCor+Rebal, MINE, and MINE+Rebal, with MINE already performing well alone and showing no additional gains from Rebalancing.

\paragraph{Qualitative disentanglement performance}
To complement the quantitative evaluation, we qualitatively assessed disentanglement using scatter plots of the task latent subspace \(z_1\), colored by the confounder labels \(y_2\) (Fig.~\ref{fig:disent-performance}). This visualization focused on the subspace-label relationship, which was identified as the most critical off-diagonal entry in the confusion matrices reported in Table~\ref{tab:disent-performance}. Since all latent spaces were two-dimensional by design, no additional projection or dimensionality reduction was required for visualization. All scatter plots were created based on the balanced test distribution. For clarity, we show only show the models from the first fold of the 5-fold cross-validation. Across datasets and methods, we observed a clear qualitative correspondence between poor quantitative disentanglement---indicated by high \(k\)NN accuracy for the \(z_1\)--\(y_2\) entry---and visible clustering of samples in \(z_1\) according to the confounder label. Conversely, methods with strong quantitative disentanglement exhibited a more homogeneous distribution of points with respect to \(y_2\), with no discernible structure in the task subspace.
Interestingly, for some methods, the learned latent representations effectively collapsed onto a single dimension, which was visible as samples lying approximately along a line in the two-dimensional latent space. This behavior was consistently observed for dCor+Rebal (Fig.~\ref{fig:disent-performance}), suggesting that the method internally emphasized a single dominant direction in \(z_1\) while suppressing confounder-related variation.

\subsection{Shortcut reliance increases with correlation strength in the training data}

\begin{figure*}[t]
    \centering
    \includegraphics[width=\linewidth]{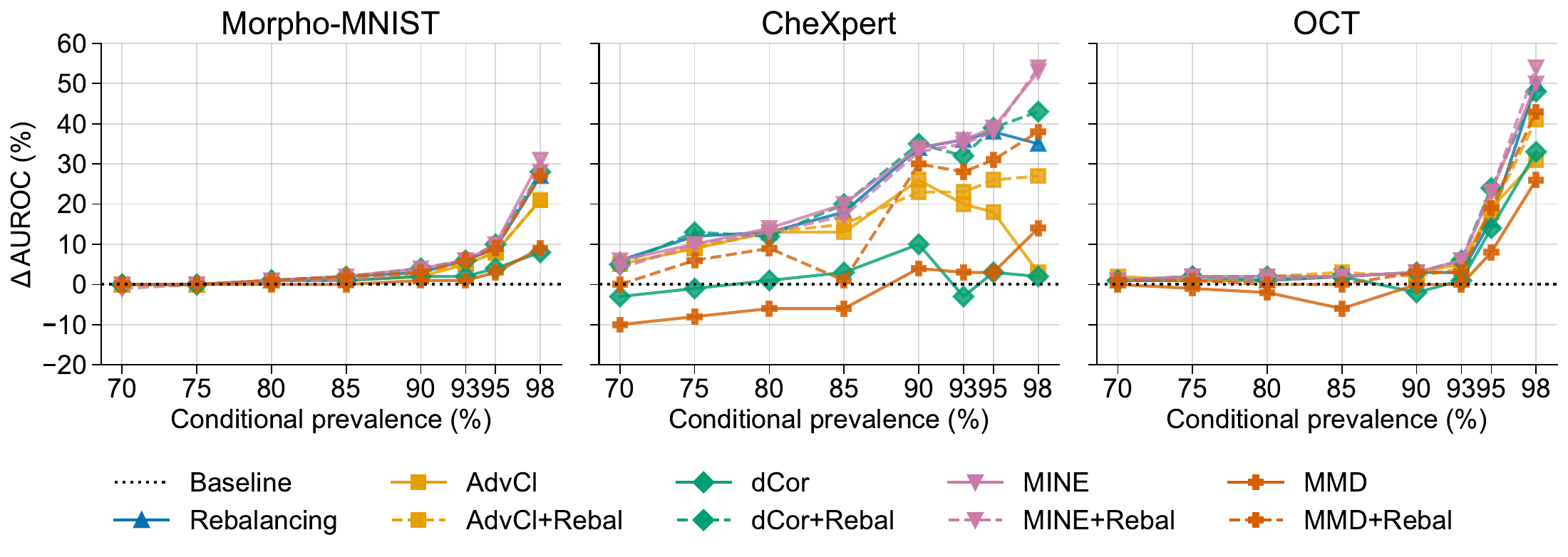}
    \caption{Relative AUROC improvement over the Baseline on the inverted test distribution (\(p(y_1=1 \mid y_2=1)=5\%\)) across different conditional prevalence levels used during training. Positive values indicate gains relative to the Baseline; variants combined with Rebalancing are shown with dashed lines.}
    \label{fig:cond-prev-ablation}
\end{figure*}

We studied the effect of varying the conditional prevalence \(p(y_1 = 1 \mid y_2 = 1)\) during training, evaluating performance on the inverted test distribution (\(p(y_1 = 1 \mid y_2 = 1) = 5\%\)). Performance is reported as \(\Delta\)AUROC relative to the baseline, with variants combined with Rebalancing shown as dashed lines (Fig.~\ref{fig:cond-prev-ablation}).

\paragraph{Artificial dataset (Morpho-MNIST)}
For Morpho-MNIST, no substantial gains over the baseline were observed for conditional prevalence up to 85\% (Fig.~\ref{fig:cond-prev-ablation}). At higher correlation strengths, improvements became more evident: at 95\%, most methods achieved \(\Delta\)AUROC of approximately 5--10\%, while at 98\%, performance differences between methods became more pronounced. The strongest improvements at 98\% (\(\Delta\)AUROC \(\approx\) 25--30\%) were observed for MINE, MINE+Rebal, dCor+Rebal, Rebalancing, and MMD+Rebal, whereas dCor and MMD achieved smaller gains (\(\approx\) 10\% \(\Delta\)AUROC).

\paragraph{Radiology (CheXpert)}
On CheXpert, relative improvements across conditional prevalences were more variable. Several methods, including Rebalancing, AdvCl, AdvCl+Rebal, dCor+Rebal, MINE, and MINE+Rebal, already showed improvements at lower prevalences (70--85\%). AdvCl without Rebalancing degraded at prevalences above 90\%, while AdvCl+Rebal remained stable up to 98\%. Across all prevalences, MINE and MINE+Rebal generally yielded the largest gains, followed by Rebalancing and dCor+Rebal, with dCor+Rebal showing clearer improvements over dCor alone.

\paragraph{Ophthalmology (OCT)}
Trends on OCT were similar to Morpho-MNIST but with larger effect sizes. Improvements emerged at higher prevalences (around 90--93\%) and reached up to 40--50\% at 98\%. MINE, MINE+Rebal, Rebalancing, and dCor+Rebal performed best overall, while AdvCl (with and without Rebalancing) and MMD+Rebal showed competitive performance at 95\% but not at 98\%.

\paragraph{Summary}
Across datasets, shortcut mitigation methods tended to yield limited benefits when conditional prevalences during training were moderate, with clearer improvements emerging under stronger correlations. The magnitude of performance gains depended not only on the mitigation strategy but also on the dataset and the confounding features present. For instance, in CheXpert, where the main task is pleural effusion and the confounding factor is sex, performance differences between methods were already noticeable at lower conditional prevalences, whereas the synthetic confounders in Morpho-MNIST and the OCT dataset required higher prevalence levels to reveal clear differences. Over all datasets, Rebalancing, MINE-based approaches, and dCor+Rebal were frequently among the better-performing configurations at high prevalence levels. Overall, these results suggest that model reliance on shortcuts---and consequently the effectiveness of shortcut mitigation---depends on the interplay between the strength of spurious correlations, the nature of the confounding features, and the mitigation strategy employed.

\subsection{Combining rebalancing with disentanglement can improve performance while maintaining computational efficiency}
In addition to evaluating the performance measures for disentanglement above, we measured the convergence for each method in epochs and minutes (Table~\ref{tab:compute-time}). We found that the baseline and the Rebalancing method consistently converged the fastest across datasets. Feature disentanglement approaches generally required longer training times. A clear outlier was MINE, which required substantially longer convergence times than all other methods. Interestingly, when combined with Rebalancing, feature disentanglement approaches typically converged faster than when applied in isolation.
\begin{table}[htbp]
    \centering
    \caption{Comparison of number of epochs and compute time in minutes until convergence.}
    \label{tab:compute-time}
    \begin{tabular}{lrr rr rrr} 
        \toprule
        & \multicolumn{2}{c}{Morpho-MNIST} 
        & \multicolumn{2}{c}{CheXpert} 
        & \multicolumn{2}{c}{OCT} \\
        \cmidrule(lr){2-3} \cmidrule(lr){4-5} \cmidrule(lr){6-7}
        Method & Epochs & Time 
               & Epochs & Time 
               & Epochs & Time \\
        \midrule
        Baseline & 97 & 1 & 9 & 6 & 11 & 5 \\
        Rebalancing & 33 & 1 & 2 & 2 & 6 & 6 \\
        AdvCl & 961 & 10 & 24 & 19 & 24 & 11 \\
        AdvCl+Rebal & 209 & 3 & 6 & 8 & 13 & 12 \\
        dCor & 536 & 6 & 19 & 14 & 24 & 11 \\
        dCor+Rebal & 206 & 3 & 2 & 3 & 11 & 10 \\
        MINE & 4451 & 501 & 132 & 52 & 280 & 76 \\
        MINE+Rebal & 505 & 103 & 31 & 23 & 126 & 63 \\
        MMD & 934 & 11 & 20 & 5 & 20 & 9 \\
        MMD+Rebal & 23 & 1 & 1 & 1 & 7 & 7 \\
        \bottomrule
    \end{tabular}
\end{table}
Given these differences, we further analyzed the trade-off between disentanglement performance and convergence time by explicitly relating disentanglement quality to computational efficiency. To quantify disentanglement, we computed a chance-normalized diagonal dominance score. Specifically, chance-level performance (50\%) was subtracted from all entries of the subspace-label accuracy matrices, followed by taking absolute values and measuring the relative dominance of diagonal over off-diagonal entries. This resulted in a score in the range \([0,1]\), where higher values indicate stronger disentanglement.
\begin{figure*}[t]
    \centering
    \includegraphics[width=\linewidth]{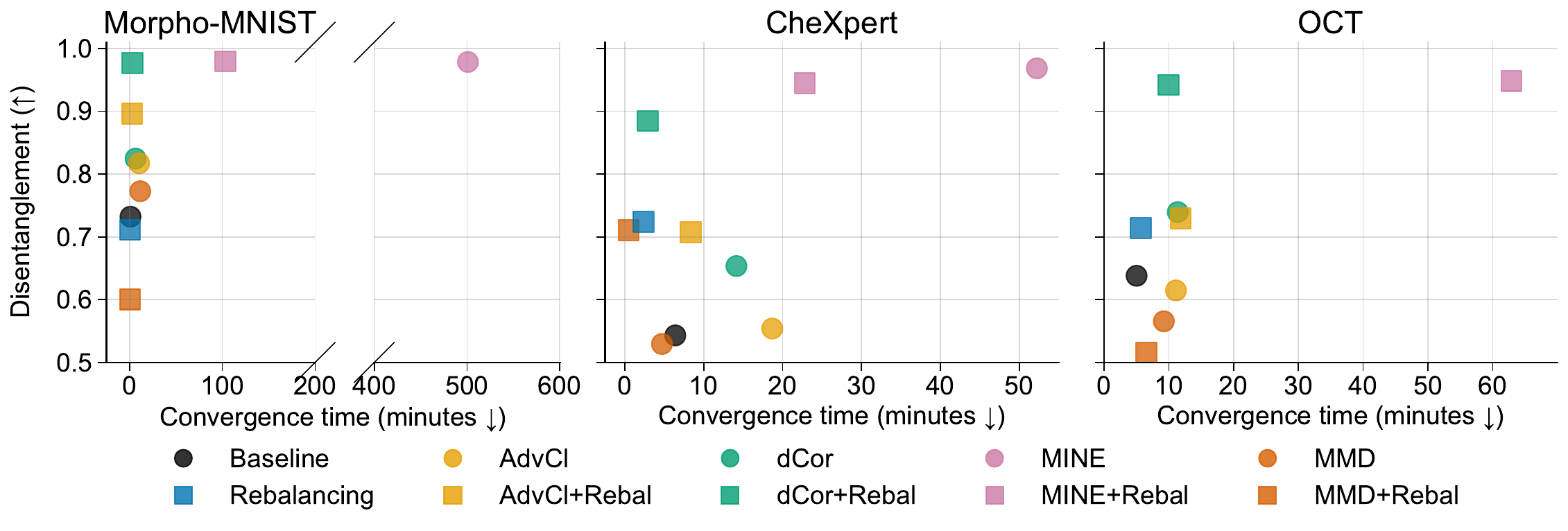}
    \caption{Disentanglement performance (diagonal dominance) of different methods as a function of convergence time in minutes.}
    \label{fig:disent_vs_time}
\end{figure*}  
Although MINE and MINE+Rebal achieved disentanglement scores near the maximum of 1.0, this came at the cost of long training times. By contrast, dCor+Rebal reached comparable disentanglement while converging significantly faster (Fig.~\ref{fig:disent_vs_time}). These results indicate that combining model-centric feature disentanglement with data-centric rebalancing can maintain strong performance while reducing computational overhead, enabling efficient training without sacrificing representation quality.

\section{Discussion}
We found that shortcut mitigation methods can improve task performance under distribution shifts. Beyond classification metrics, we observed clear differences in feature disentanglement in latent spaces, and we found that shortcut reliance increased with the strength of spurious correlations in the training data. We also found that combining data rebalancing with disentanglement improves performance in some cases while maintaining computational efficiency. Overall, MINE and dCor+Rebal consistently performed best, with dCor+Rebal being more computationally efficient.

Our results point to several limitations and practical considerations when using feature disentanglement to mitigate shortcut learning in medical imaging. Our findings show that MMD~\cite{gretton2012kernel}, particularly when combined with rebalancing, generally showed weaker disentanglement than the other methods. This may be due to the sensitivity of kernel-based measures to the feature distributions, which requires further study. Visualizations of the latent spaces showed that, for some methods, the task-related subspace collapsed to a single dominant dimension, with samples lying close to a line. This indicates that only a small part of the latent space was effectively used. We used a setting with 95\% conditional prevalence to ensure consistent comparison across methods. While this represents a strong level of confounding, prior work indicates that similar correlations can occur in real clinical data. For example, Degrave et al.~\cite{degrave2021ai} showed that in a widely used COVID-19 chest radiograph dataset, almost all positive cases came from a single data source, while negative cases came from different hospitals and time periods. As a result, models learned to rely on source-related artifacts rather than disease-related features, such as laterality markers, annotations, and image border patterns introduced during image acquisition. This example illustrates that when datasets are combined from multiple sources, as is common in medical imaging, strong correlations between confounders and labels can naturally arise. This supports the use of high conditional prevalences when evaluating shortcut mitigation methods. At the same time, controlling a single confounder does not rule out the presence of other shortcuts. As noted by Degrave et al.~\cite{degrave2021ai}, models may rely on multiple, redundant shortcuts and can shift to alternative spurious cues once one shortcut is mitigated. The goal of this work was not to identify or remove all possible shortcuts, but to systematically study how well different disentanglement methods can mitigate a known shortcut under a controlled experimental setting.

Several directions could further improve shortcut mitigation in medical imaging. First, evaluating disentanglement methods on larger and more complex datasets, such as NAKO or UK Biobank, would help assess whether large-scale data or foundation models naturally reduce shortcut learning. Current evidence suggests that this is not guaranteed, as merging data from multiple hospitals can introduce strong confounder-label correlations even at scale~\cite{zhang2025adversarial}. Prior work has shown that careful dataset curation, which explicitly avoids such confounders, can improve out-of-distribution performance, even when it results in smaller datasets~\cite{degrave2021ai}. Second, extending the analysis to settings with multiple confounders and their interactions would better reflect real clinical scenarios, where several confounding factors often coexist~\cite{fay2025mimm}. Third, using higher-dimensional latent spaces may allow models to capture more complex features than the two-dimensional subspaces studied here. Fourth, combining complementary disentanglement strategies, such as adversarial learning and distance correlation, may help to mitigate different aspects of shortcut learning. Finally, to strengthen clinical relevance, future work could explore whether disentangled representations improve interpretability, and how these methods perform in longitudinal or continual learning settings with evolving data distributions.

\section*{Acknowledgments}
This project was supported by the Hertie Foundation and by the Deutsche Forschungsgemeinschaft under Germany's Excellence Strategy with the Excellence Cluster 2064 ``Machine Learning---New Perspectives for Science'', project number 390727645. This research utilized compute resources at the Tübingen Machine Learning Cloud, INST 37/1057-1 FUGG. The project was additionally funded by the Hertie Foundation. PB is a member of the Else Kröner Medical Scientist Kolleg ``ClinbrAIn: Artificial Intelligence for Clinical Brain Research''. The authors thank the International Max Planck Research School for Intelligent Systems (IMPRS-IS) for supporting SM.

\bibliographystyle{plainnat}
\bibliography{references}

\end{document}